\documentclass[runningheads]{llncs}
\usepackage{graphicx}
\usepackage[colorlinks,linkcolor=red, anchorcolor=blue, citecolor=blue, urlcolor=magenta]{hyperref}
\usepackage{tikz}
\usepackage{comment}
\usepackage{amsmath,amssymb} 
\usepackage{color}

\usepackage[accsupp]{axessibility}  

\usepackage{graphicx}
\usepackage{amsmath}
\usepackage{amssymb}
\usepackage{booktabs}
\usepackage{multirow}
\usepackage{tabulary}
\usepackage{colortbl}
\usepackage{booktabs}
\usepackage{amsmath}
\usepackage{algorithmic,algorithm}

\usepackage{listings}

\begin{document}
\pagestyle{headings}
\mainmatter
\def\ECCVSubNumber{4481}  

\title{Dual Adaptive Transformations for Weakly Supervised Point Cloud Segmentation} 

\titlerunning{Dual Adaptive Transformations}
%
\author{Zhonghua Wu\inst{1,2} \and
Yicheng Wu\inst{3} \and
Guosheng Lin\thanks{Corresponding author: G. Lin (e-mail: gslin@ntu.edu.sg)}\inst{1,2} \and
Jianfei Cai\inst{2,3}\and
Chen Qian\inst{4}}
\authorrunning{Z. Wu et al.}
%
\institute{S-lab, Nanyang Technological University \and
School of Computer Science and Engineering, Nanyang Technological University \and
Department of Data Science and AI, Monash University \and
SenseTime Research\\
\email{zhonghua001@e.ntu.edu.sg}}
\maketitle

\begin{abstract}
Weakly supervised point cloud segmentation, i.e. semantically segmenting a point cloud with only a few labeled points in the whole 3D scene, is highly desirable due to the heavy burden of collecting abundant dense annotations for the model training. However, existing methods remain challenging to accurately segment 3D point clouds since limited annotated data may lead to insufficient guidance for label propagation to unlabeled data. Considering the smoothness-based methods have achieved promising progress, in this paper, we advocate applying the consistency constraint under various perturbations to effectively regularize unlabeled 3D points. 
Specifically, we propose a novel DAT (\textbf{D}ual \textbf{A}daptive \textbf{T}ransformations) model for weakly supervised point cloud segmentation, where the dual adaptive transformations are performed via an adversarial strategy at both point-level and region-level, aiming at enforcing the local and structural smoothness constraints on 3D point clouds.
We evaluate our proposed DAT model with two popular backbones on the large-scale S3DIS and ScanNet-V2 datasets. Extensive experiments demonstrate that our model can effectively leverage the unlabeled 3D points and achieve significant performance gains on both datasets, setting new state-of-the-art performance for weakly supervised point cloud segmentation. 
\end{abstract}

\section{Introduction}
\label{sec:intro}
Recently, the deep learning (DL)-based methods have achieved significant performance gains for the point cloud segmentation task, which is a fundamental and critical step to understand realistic scenes \cite{jaritz2019multi} and analyze 3D geometric data \cite{zhou2018voxelnet}. However, it is extremely costly and labor-consuming to collect abundant dense annotations of 3D point clouds for model training. Thus, it is highly desirable to develop effective algorithms that can well segment point cloud data with only weak annotations of point clouds. 

\begin{figure}[t]
\centering
    \includegraphics[width=\linewidth]{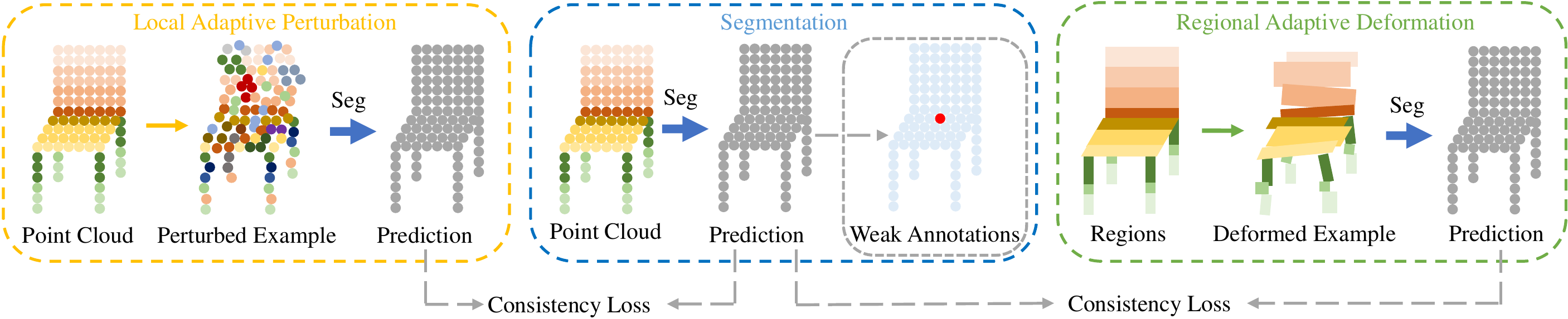}
    \caption{Illustration of the proposed Dual Adaptive Transformation (DAT) model. We encourage DAT to produce consistent predictions under local and regional adaptive transformations. Note that, there are only few labeled points inside the whole scene to train our model. During testing, only segmentation module (blue) is used to generate the segmentation prediction.}
    \label{fig:overview}
\end{figure}

For semantic image segmentation tasks, there are different types of weak annotations including image-level labels~\cite{oh2017exploiting,wang2018weakly,zhang2020splitting}, scribbles~\cite{lin2016scribblesup,wu2019keypoint}, or partially labeled samples~\cite{8935407,liu2021few,wu2021learning}.
For the point cloud segmentation task, following the recent work~\cite{liu2021one}, we consider partially labeled samples as weak annotations for the model training, i.e., only a few sparse points inside the whole scene are labeled and all other points are unlabeled. The latest model 1T1C~\cite{liu2021one} attempts to train a segmentation model with limited labeled points and then propagate the labels to the unlabeled points as the pseudo labels for iteratively refining the model. However, such a training strategy is time-consuming and is often affected by unreliable pseudo labels, resulting in sub-optimal segmentation performance. Here, we hypothesize that the weakly supervised segmentation performance can be further improved by adding more constraints on the unlabeled 3D points.

To exploit the unlabeled data, the consistency-based learning methods have shown promising progress in natural image classification and segmentation. For example, \cite{sohn2020fixmatch,berthelot2019mixmatch,french2019semi} encouraged the model to produce invariant results under various strong data augmentations. However, it is non-trivial to apply these image-based strong data augmentation techniques to point cloud processing, and point cloud-specific augmentations are still under early exploration \cite{chen2020pointmixup,li2020pointaugment}. This motivates us to investigate an effective transformation method to leverage large amounts of unlabeled 3D points by applying sufficient smoothness constraints for weakly supervised point cloud segmentation. 

Specifically, in this paper, we propose a \textbf{Dual Adaptive Transformation (DAT)} model, where we encourage consistent predictions between original and local/regional adaptively transformed point clouds data.
As shown in Figure~\ref{fig:overview}, we first design a \textit{Local Adaptive Perturbation (LAP)} module that computes the adaptive perturbations for both point coordinates and their associate features. Meanwhile, considering the feature distributions are quite different between different classes, we further embed the class-aware information into the LAP module to generate class-aware adaptive feature perturbations. Then, to capture more structural information in point clouds, we further introduce a \textit{Regional Adaptive Deformation (RAD)} module to apply adaptive deformations on the pre-defined super-points, which enforces the consistency constraints at the region level.

We evaluate our DAT model with two popular backbones on the large-scale S3DIS dataset \cite{armeni20163d} and ScanNet-v2 dataset \cite{dai2017scannet}. Via effectively leveraging the unlabeled point clouds, our DAT model is able to segment point cloud data with very few annotations, setting new state-of-the-art (SOTA) performance for the weakly supervised point cloud segmentation task. For example, on S3DIS dataset \cite{armeni20163d}, the DAT model outperforms the previous SOTA model 1T1C \cite{liu2021one} by 6.5\% under the ``One Thing One Click'' annotation setting. Note that our proposed strategy can be easily combined with other frameworks. For instance, based on our design, the segmentation performance of 1T1C~\cite{liu2021one} model can be further improved by 2.9\%/3.0\% on the ScanNet-v2 test/validation set \cite{dai2017scannet}, respectively.

Overall, our main contributions are three-fold:
\begin{itemize}
\item We propose a novel Dual Adaptive Transformation (DAT) model for weakly supervised point cloud segmentation, with the key insight that applying the consistency constraint under local and regional adaptive transformations can effectively leverage a large amount of unlabeled 3D points and facilitate a better model training.
\item We introduce the Local Adaptive Perturbation (LAP) module, where we inject the adaptive perturbations to point coordinates and the associate feature inputs separately. Meanwhile, we embed the information of the class-aware point feature distribution into the generation of the local adaptive feature perturbations, which leads to better performance.
\item We introduce the Regional Adaptive Deformation (RAD) module, where we generate structural adaptive deformations at the region-level, i.e. adaptive deformations such as shifting, scaling, and rotation for the superpoint regions. Such regional deformations introduce another level of the consistency constraint, which is a complement to LAP.
\end{itemize}

\section{Related Work}
\subsection{Deep Learning on Point Clouds}
DL-based methods have achieved great progress to process point cloud data. For example, PointNet model \cite{qi2017pointnet} used permutation-invariant operators such as pooling layers to aggregate the features from all points. Then, PointNet++ model \cite{qi2017pointnet++} further designed a hierarchical spatial structure to extract local geometric features. Furthermore, the graph-based methods \cite{wang2019dynamic,li2019deepgcns,landrieu2018large} built a graph for all points and applied the message passing mechanism on the graph. For instance, DGCNN \cite{wang2019dynamic} used a kNN graph to perform graph convolutions. To capture contextual relationships, SPG \cite{landrieu2018large} constructed a graph on the sub-regions, i.e. the superpoints. DeepGCNs \cite{li2019deepgcns} explored the depth information in graph convolutional networks. Afterwards, \cite{xu2018spidercnn,su2017learning,thomas2019kpconv,mao2019interpolated} further improved the performance by directly applying continuous convolutions on the points without any quantization. SpiderCNN \cite{xu2018spidercnn} used polynomial functions to generate the kernel weights and the spherical convolution \cite{su2017learning} was used to address the 3D rotation equivariance problem in Spherical CNN. KPConv \cite{thomas2019kpconv} constructed the kernel weights based on the input coordinates and achieved good performance. Similarly, InterpCNN \cite{mao2019interpolated} interpolated point-wise kernel weights by utilizing the coordinate information. Different from point convolution networks, the voxel-based methods \cite{choy20194d} firstly quantized all the points and map the points to the regular voxels and then applied 3D convolutions on the regular voxels to obtain point features.

In this paper, we adopt the point-based KPConv model \cite{thomas2019kpconv} as our backbone, where the model is trained via encouraging the dual adaptive transformation consistency for weakly supervised point cloud segmentation. Furthermore, in Section~\ref{generalization}, we also extend our method to the voxel-based framework MinkowiNet \cite{choy20194d} so as to demonstrate the generalization ability of our training strategy.

\subsection{Weakly Supervised Point Cloud Segmentation}
There are some DL-based methods being proposed recently for the weakly supervised point cloud segmentation task~\cite{wang2021new,meng2021towards,tao2020seggroup,gao2020we,zhu2021weakly,li2021snapshotnet,deng2021superpoint,hu2021sqn,zhang2021perturbed,hamdi2020advpc,xiang2019generating,nekrasov2021mix3d}. For example, Wang et al. \cite{wang2020weakly} proposed to generate point cloud segmentation labels by back-projecting 2D image annotations to 3D spaces. However, annotating large-scale image semantic segmentation datasets is extremely labor-consuming. To reduce the labeling costs, Wei et al. \cite{wei2020multi} used the Class Activation Map (CAM)~\cite{zhou2016learning,wu2020exploring} to generate pseudo segmentation masks with sub-cloud level annotations. However, its performance is limited due to the lack of localization information in labels. To address the issue, Xu et al. \cite{xu2020weakly} further labeled 10\% points in the whole point cloud, which is able to achieve a good performance comparable to the fully-supervised references. Then, the 1T1C method \cite{liu2021one} under the ``One Thing One Click'' setting was introduced to tackle this task, which uses fewer labeled points, i.e. only labeling one point per thing in each scene.

Here, we follow the 1T1C method~\cite{liu2021one} to conduct experiments. Different from the iterative refinement mechanism used in 1T1C which brings in significant computational cost, we propose an end-to-end training strategy to train a model in the identical weakly supervised manner while without the need for any iterative refinement.

\subsection{Consistency-based Semi-supervised Learning}
Our work is closely related to the consistency based semi-supervised learning (SSL)~\cite{wu2022exploring,wu2022mutual}, where the basic idea is to leverage the unlabeled data based on the smoothness assumptions, i.e. deep models under various small perturbations or augmentations should output consistent results. For example, Bortsova et al.~\cite{bortsova2019semi} enforced the model to produce invariant predictions for unlabeled images under different transformations. For semi-supervised image classification task, the VAT model \cite{miyato2018virtual} designed an adversarial perturbation and then encouraged the consistency between the original data and its adversarial one. Temporal ensembling~\cite{laine2016temporal} and mean teacher~\cite{tarvainen2017mean} generated similar distributions for the perturbed inputs. Meanwhile, the mutual learning strategy has been studied for semi-supervised learning \cite{zhang2018deep,wu2021semi}. For instance, the dual-student model \cite{ke2019dual} enforced two sub-networks learn from each other via constraining the consistent predictions. FixMatch \cite{sohn2020fixmatch} further explicitly generated the pseudo labels from the data with weak augmentations and used them to guide the prediction from the strongly augmented samples.

Motivated by the consistency-based semi-supervised learning methods which encourage the model to produce consistent results under various perturbations, we propose the DAT model for the weakly supervised point cloud segmentation task with two major novel designs, i.e. the LAP and RAD modules.

\section{Methods}

\begin{figure*}[t]
\centering
    \includegraphics[width=\linewidth]{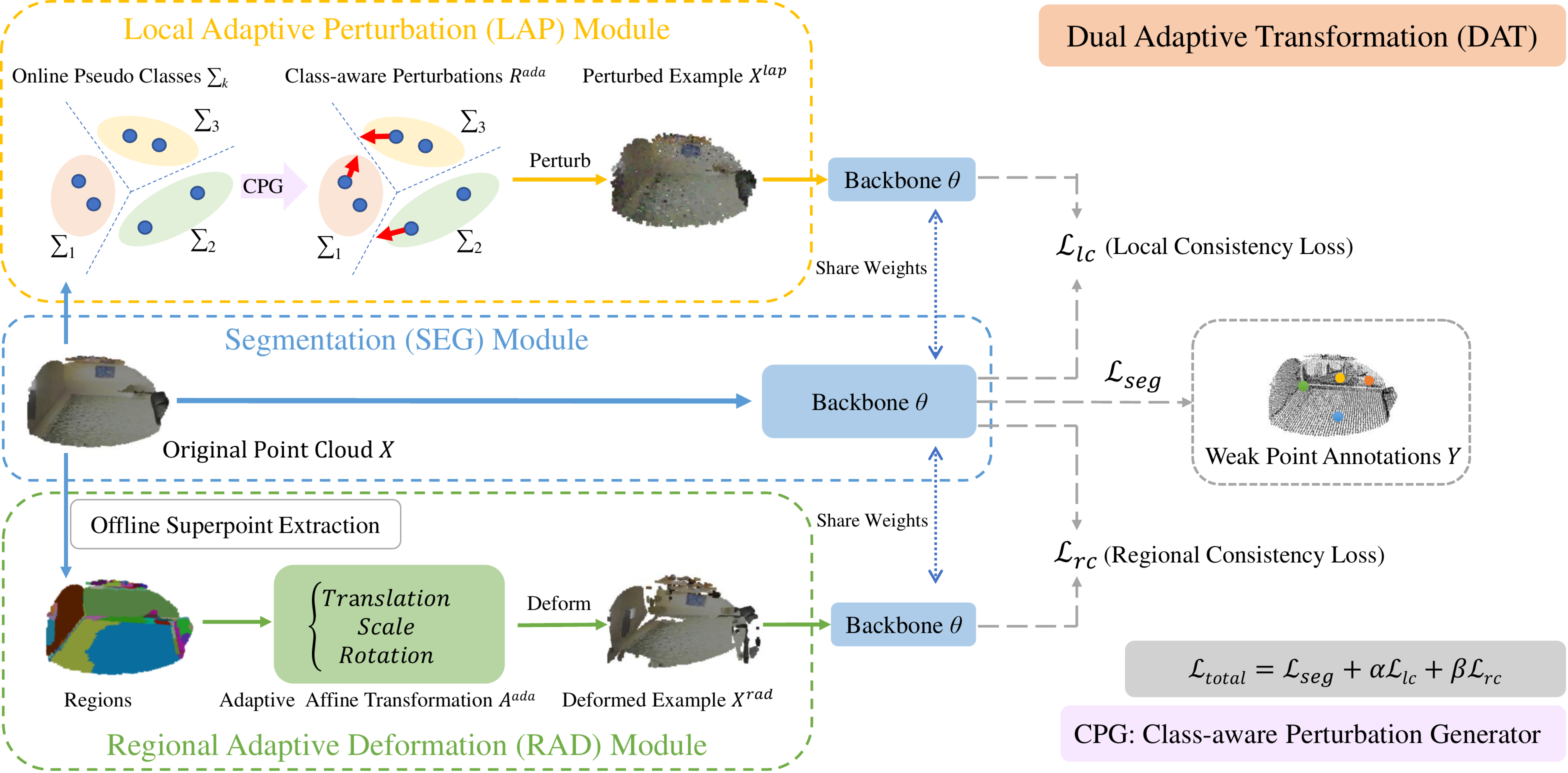}
    \caption{Overall pipeline of our proposed Dual Adaptive Transformation (DAT) model, which consists of three main modules: the segmentation (SEG) module (blue), the Local Adaptive Perturbation (LAP) module (yellow), and the Regional Adaptive Deformation (RAD) Module (green). SEG module adopts KPConv backbone to train the model with few labeled points. LAP module is to generate class-aware perturbed examples on each point. RAD module generates structural deformed data by applying the adaptive affine transformations on each region. Note that, during testing, we only employ SEG module to process point cloud data.}
    \label{fig:pipeline}
\end{figure*}

Figure~\ref{fig:pipeline} gives an overview of the proposed Dual Adaptive Transformation (DAT) model, which consists of three main modules: the class-aware Local Adaptive Perturbation (LAP) module, the Regional Adaptive Deformation (RAD) module, and the original SEGmentation (SEG) module. LAP contains a novel Class-aware Perturbation Generator (CPG) to produce semantic perturbations at the point level. RAD generates structural augmented examples by applying various deformations at the region level. SEG contains a conventional point cloud segmentation backbone.

\subsection{Segmentation Module}
We first define a set of notations for the weakly supervised point cloud segmentation task. Specifically, consider a set of points $X = [C, F] \in \mathbb{R}^{N \times 3+D_f}$ with point coordinates $C \in \mathbb{R}^{N \times 3}$ and the corresponding features $F \in \mathbb{R}^{N \times D_f}$ as the model input, and denote $Y \in \mathbb{R}^{N \times 1}$ as the groundtruth label, which is a very sparse one with only $M$ known entries, $M << N$. 
The output of SEG module is the predicted segmentation mask $\hat{Y}$. The segmentation module aims to train the backbone model with few labeled points in $Y$. Here, we adopt a popular segmentation framework KPConv \cite{thomas2019kpconv} as our backbone. With the kernel parameters denoted as $\theta$, the model prediction is given by $p(\hat{y}_i|c_i, f_i; \theta)$, $i \in \{1, ..., N\}$, where $c_i$ and $f_i$ are respectively the point coordinates and features of point $x_i$. We train the segmentation module by applying a cross-entropy loss $\mathcal{L}_{seg}$ on the few labels in $Y$ and the corresponding predictions in $\hat{Y}$.

\subsection{Local Adaptive Perturbation Module}
We design a Local Adaptive Perturbation (LAP) module to generate perturbed examples $X^{lap}$ by applying the adaptive perturbations on the point coordinates and the corresponding features. In particular, the input to LAP is the point cloud $X$ and the output is the perturbed examples $X^{lap}$ with the injection of the adaptive perturbations $R^{ada}$. Inspired by VAT~\cite{miyato2018virtual}, which is proposed for semi-supervised image classification, to achieve local distributional smoothness (LDS) as a smoothness constrain to regularize unlabeled data, we encourage our model to generate consistent outputs between each input point $x \in X$ and its perturbed version $x+r^{ada}$ , where $r^{ada} \in R^{ada}$ is the corresponding adaptive perturbation:
\begin{equation}\label{lds}
    \mathcal{LDS}(x; \theta) = D\left[p(\hat{y}|x; \theta),p(\hat{y}|x+r^{ada};\theta)\right].
\end{equation}
Here $D$ is a non-negative loss function to measure the divergence between $x$ and $x+r^{ada}$. 
Then, we compute $r^{ada}$ by estimating a gradient $g$ of $\mathcal{LDS}$ with a random input vector $d$ as
\begin{equation}
\label{eq}
\begin{aligned}
&g = \nabla_{R} D\left[p(\hat{y}|x, \theta), p(\hat{y}|x+r, \theta)\right]\Big|_{r=\xi d} \\
&r^{ada} = \epsilon \times g / \|g\|_2,
\end{aligned}
\end{equation}
where $\xi$ and $\epsilon$ are two hyper-parameters to control the magnitude of the perturbation, and $g$ can be efficiently computed by applying the back-propagation on the network.

Considering the input point coordinates and features are two different types of inputs, we generate their perturbations separately. In other words, for an input point $x$ consisting of its coordinates $c$ and features $f$, we generate the adaptive perturbation data $c+r^{ada}_c$ and $f+r^{ada}_f$ with the initial random unit vectors $d_{c}$ and $d_{f}$, respectively.

\textbf{Class-aware Perturbation Generator.}
Note that many existing perturbation based semi-supervised image classification methods~\cite{miyato2018virtual} usually generate the initial perturbations $d$ through sampling them from an iid Gaussian distribution. However, in the point cloud segmentation task, directly applying this to generate $d_{f}$ might not be optimal. This is because, for different classes, their input point feature distributions are quite different across different dimensions. A class-agnostic iid Gaussian sampling might generate unrealistic perturbations.

Therefore, we propose a Class-aware Perturbation Generator (CPG) to obtain $d_{f}$ for each point. Specifically, in each training iteration, we generate the pseudo labels $\hat{y}$ for all the points with the current model parameter $\hat{\theta}$, where $\hat{y} \in \{1, ..., K_c\}$ with $K_c$ being the number of classes.
Based on that, we establish a zero-mean multivariate normal distribution $\mathbb{N}(0, \sum_k)$. Here $\sum_{k}$ is the class-conditional covariance matrix estimated from all the input point features (e.g. rgbh for KPConv) that belong to the pseudo-class $k$. Afterward, we update the covariance matrix in an online manner~\cite{wang2021regularizing} with the statistics of the features from each mini-batch. In this way, at each iteration, $d_{f}$ is then generated by sampling from the up-to-date class-aware multivariate Gaussian distribution. 
For $d_{c}$, we adopt the conventional way i.e. sampling the initial input vectors from an iid Gaussian distribution, since the point clouds are unordered and the individual coordinates alone are not closely related to the class of the points. This is also observed in the PointNet model~\cite{qi2017pointnet}.

\begin{figure}[t]
\centering
    \includegraphics[width=0.8\linewidth]{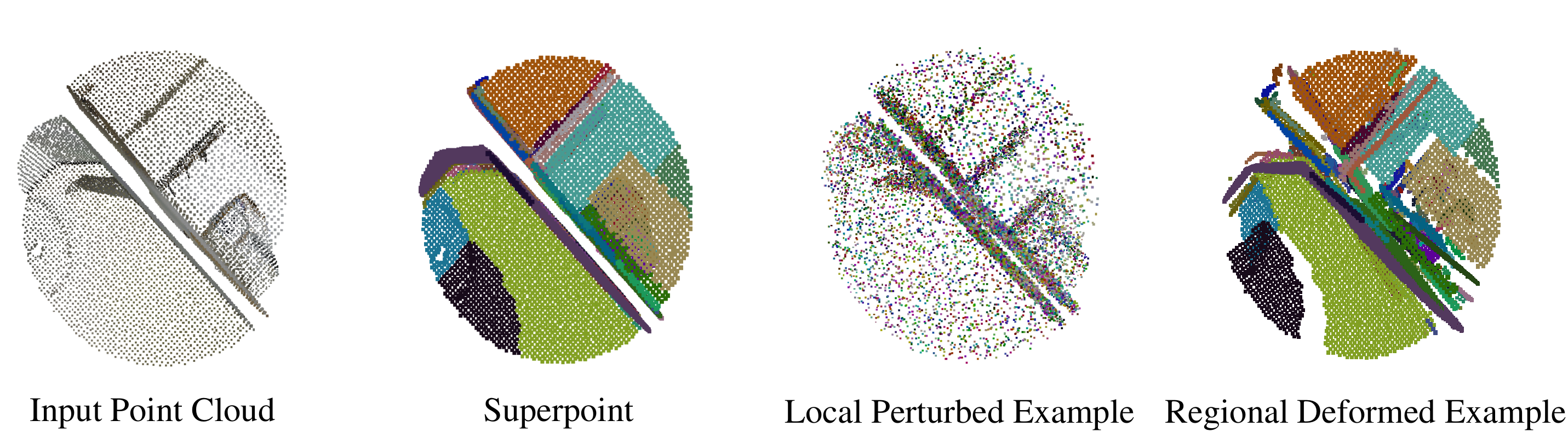}
    \caption{Visual results for the superpoint estimation and the generated dual adaptive transformed examples during the training stage.}
    \label{fig:vis_ablation}
\end{figure}

With the generated $d_c$ and $d_{f}$, our LDS loss for point clouds now becomes: 
\begin{equation}
\begin{aligned}
\mathcal{LDS}(x; \theta) &= D\left[p(\hat{y}|c, f; \theta), p(\hat{y}|c+\xi_{c} d_{c}, f+\xi_{f} d_{f}; \theta)\right] \\
g_{c} &= \nabla_{\xi_{c} d_{c}} \mathcal{LDS}(x, \theta) \\
g_{f} &= \nabla_{\xi_{f} d_{f}} \mathcal{LDS}(x, \theta),
\end{aligned}
\end{equation}
where we use the Kullback–Leibler divergence (KL-div) for $D$. Finally, we obtain the $r_c^{ada}$ and $r_f^{ada}$ by 
\begin{equation}
\begin{aligned}
&r_{c}^{\rm ada} = \epsilon_{c} g_{c}/\|g_{c}\|_2 \\
&r_{f}^{\rm ada} = \epsilon_{f} g_{f}/\|g_{f}\|_2.
\end{aligned}
\end{equation}
In this way, the perturbed examples $X^{lap}$ is obtained by point-wise adding the perturbations $r_c^{\rm ada}$ and $r_f^{\rm ada}$ on the coordinates $c$ and the features $f$, respectively. One example is visualized in the third column of Figure~\ref{fig:vis_ablation}.

\subsection{Regional Adaptive Deformation Module}
In addition to the local adaptive perturbations, considering point clouds often contain various structural local deformations such as region shift, rotation, and scaling, we further design a regional adaptive deformation (RAD) module to generate structural local deformations.
RAD module takes point cloud $X$ as input and outputs region-level augmented examples $X^{rad}$ by deforming each region with adaptive affine transformations $A^{ada}$. As shown in Figure~\ref{fig:pipeline}, we firstly over-segment point cloud $X$ into a set of superpoints $S_i, i \in \{1, ..., K_s\}$ via \cite{landrieu2018large,dai2017scannet}. For each superpoint $S_i$, we generate the adaptive deformed example $S_i^{ada}$. Combing all $S_i^{ada},  i \in \{1, ..., K_s\}$, we obtain $X^{rad}$.

For each superpoint $S_i$, we firstly generate the initial affine transformation matrices $A_{i,j}$, whose parameters are randomly sampled from an iid Gaussian distribution. Then, we deform each superpoint as
\begin{equation}
    S_i^{int} = S_i \cdot \prod_{j=1}^{K_a} \xi_{A} A_{i,j},
\end{equation}
where $A_{i,j}, j \in \{1, ..., K_a\}$, corresponds to the $j$-th type of deformations. Combining all $S_i^{int},  i \in \{1, ..., K_s\}$, we obtain the initial deformed point cloud $X^{int}$. The $\mathcal{LDS}$ loss becomes
\begin{equation}
\begin{aligned}
\mathcal{LDS}(X; \theta) &= D\left[p(\hat{y}|x; \theta), p(\hat{y}|x^{int}; \theta)\right] \\
g_{A_{i,j}} &= \nabla_{\xi_{A} A_{i,j}} \mathcal{LDS}(x; \theta) .
\end{aligned}
\end{equation}
Then, we obtain the $A^{ada}_{i,j}$ by
\begin{equation}
A^{ada}_{i,j} = \epsilon_{A} g_{A_{i,j}}/\|g_{A_{i,j}}\|_2.
\end{equation}
Finally, the regional deformed examples $X^{rad}$ is obtained by combining all the deformed superpoints $S_i^{ada}$, which is computed as
\begin{equation}
    S_{i}^{ada} = S_i * \prod_{j=1}^{K_a} A_{i, j}^{ada}.
\end{equation}
Specifically, we use the following three types of affine transformations: translation, scale and rotation.

\begin{algorithm}[t]
\caption{\label{algo:vat_loss} Generating adaptive transformed examples (LAP/RAD)}
    \textbf{Input:}{ Training Point Cloud $X$}\\
    \textbf{Output:}{ Local perturbed examples $X^{lap}$/ Regional deformed examples $X^{rad}$}
\begin{enumerate}

\item Generate initial $R$/$A$ for initial transformation.   
\item Compute the gradient of $D$ with respective to $R$/$A$ \\
$g_R/g_A \leftarrow \nabla_{r} D\left[p(\hat{y}|x; \theta), p(\hat{y}|x \odot R ; \theta)\right]\Big|_{r=\xi (R/A)}$
where $\odot$: $\oplus$/$\otimes$ for LAP/RAD, respectively.
\item Normalize the gradient to generate adaptive perturbations $R^{ada}$/$A^{ada}$.\\
$R^{ada} \leftarrow \epsilon \cdot g_R/\|g_R\|_2$ or $A^{ada} \leftarrow \epsilon \cdot g_A/\|g_A\|_2$
\item Generate the adversarial examples $X^{lap}$ / $X^{rad}$ by injecting the $R^{ada}$ / $A^{ada}$ to Point Cloud $X$.
\end{enumerate}
\end{algorithm}

One RAD example is given in the fourth column of Figure~\ref{fig:vis_ablation}. Algorithm~\ref{algo:vat_loss} summarizes the process of generating the adversarial examples under both LAP and RAD.

\subsection{Training Losses}
The overall training loss can be written as 
\begin{equation}
\mathcal{L}_{total} =\mathcal{L}_{seg} + \alpha \mathcal{L}_{lc} + \beta \mathcal{L}_{rc}
\end{equation}
where $\mathcal{L}_{seg}$, $\mathcal{L}_{lc}$ and $\mathcal{L}_{rc}$ are \textit{Segmentation Loss}, \textit{Local Consistency Loss} and \textit{Regional Consistency Loss}, respectively, and $\alpha$ and $\beta$ are trade-off weights, both set as 2 to balance the losses. {Segmentation Loss} $\mathcal{L}_{seg}$ is to guide the segmentation prediction with the limited annotations in $Y$. Specifically, we follow the KPConv~\cite{thomas2019kpconv} by using the cross entropy loss for $\mathcal{L}_{seg}$ to train the segmentation prediction $\hat{Y}$. 
{Local Consistency Loss} $\mathcal{L}_{lc}$ encourages the consistency and penalizes the prediction difference between the original point cloud $X$ and the local perturbed examples $X^{lap}$. {Regional Consistency Loss} $\mathcal{L}_{rc}$ ensures the consistency between $X$ and its regional deformed examples $X^{rad}$. $\mathcal{L}_{pc}$ and $\mathcal{L}_{rc}$ are defined as 
\begin{equation}
\begin{aligned}
&\mathcal{L}_{pc} = D\left[p(\hat{y}|x; \theta), p(\hat{y}|x^{lap}; \theta)\right] \\
&\mathcal{L}_{rc} = D\left[p(\hat{y}|x; \theta), p(\hat{y}|x^{rad}; \theta)\right]
\end{aligned}
\end{equation}
where $D$ is the KL-div loss. 

\section{Experiments and Results}
\subsection{Implementation Details}
\noindent\textbf{Datasets.} Following the 1T1C~\cite{liu2021one} model, we conduct experiments on two large-scale point cloud datasets - the S3DIS \cite{armeni20163d} and ScanNet-v2 \cite{dai2017scannet}. The S3DIS dataset consists of 3D scans of 271 rooms with 13 categories belonging to 6 areas. For fair comparisons, we train the segmentation model on Area 1, 2, 3, 4, 6 and test on Area 5 as \cite{liu2021one}. The ScanNet-v2 dataset contains 1201, 312, and 100 3D scans for training, validation, and testing, respectively.

\noindent\textbf{Weak Annotation Scheme.}
For fair comparisons, on the S3DIS dataset, we label the data under the ``One Thing One Click'' (OTOC) setting as in 1T1C \cite{liu2021one}. We randomly select a point in each object with the identical probability as the labeled points. Therefore, only 0.02\% of points have annotations inside the whole point cloud. On the ScanNet-v2 dataset, we evaluate our DAT model on the ``3D Semantic label with Limited Annotations'' benchmark \cite{dai2017scannet}. In this benchmark, only 20 points are labeled in each room scene.

\noindent\textbf{Experiment Setting.}
If there is no special declaration, we implement our proposed DAT training method based on the KPConv \textit{rigid} model. We use SGD to train the model with learning rate of 0.01 and batch size of 2. Following 1T1C \cite{liu2021one}, we use the geometrical partition results~\cite{landrieu2018large} and mesh segment results~\cite{dai2017scannet} as the superpoints for S3DIS and ScanNet-v2 datasets, respectively. We set the hyper-parameters $\xi_{c} = 10$, $\xi_{f} = 0.1$, $\xi_{A}=0.1$, $\epsilon_{c} = 1$, $\epsilon_{f} = 0.05$, $\epsilon_{A} = 0.05$. During the model training, to reduce the GPU memory consumption, we employ the segmentation loss $\mathcal{L}_{seg}$ at all iterations and randomly apply local consistency loss $\mathcal{L}_{lc}$ or regional consistency loss $\mathcal{L}_{rc}$ with an equal
probability of 0.5 to train our model. All of our experiments are conducted on a single NVIDIA RTX 3090 GPU with PyTorch 1.7.0 and CUDA 11.0.

\subsection{Evaluations on S3DIS dataset}

\subsubsection{Comparing with State-of-the-art Methods.}
\label{generalization}

\begin{table}[t]
\caption{Comparison of our DAT with several existing methods on the S3DIS Area-5 set. Note that, we report the performance as final results based on the KPConv \cite{thomas2019kpconv} backbone.}
\centering

\resizebox{0.5\linewidth}{!}{
\begin{tabular}{c|c|c}
\toprule[1.5pt]
Method                                                               & Supervision (\%) & mIoU (\%)                      \\ \midrule
PointNet \cite{qi2017pointnet}                 & 100\%            & 41.1                           \\
PointCNN \cite{li2018pointcnn}                 & 100\%            & 57.3                           \\
Xu et al. \cite{xu2020weakly}                  & 0.2\%            & 44.5                           \\
Xu et al. \cite{xu2020weakly}                  & 10\%             & 48.0                           \\
GPFN \cite{wang2020weakly}                     & 16.7\% 2D        & 50.8                           \\
GPFN \cite{wang2020weakly}                     & 100\% 2D         & 52.5                           \\ \midrule \midrule
1T1C \cite{liu2021one}                         & 0.02\% (OTOC)    & 50.1                           \\
1T1C \cite{liu2021one}                         & 0.06\% (OTTC)    & 55.3                           \\ 
\midrule
Our DAT                                                          & 0.02\% (OTOC)    & \textbf{56.5} \\
Our DAT                                                          & 0.06\% (OTTC)    & \textbf{58.5} \\ \midrule 
Our Upper Bound                                                      & 100\%            & 65.4                           \\ \bottomrule[1.5pt]
\end{tabular}
}
\label{tab:s3dis-compare-sota}
\end{table}

\begin{table}[t]
\centering
\caption{Comparison of our DAT with its variant methods with the KPConv framework. Note that, all experiments are conducted under the OTOC setting on the S3DIS dataset}
\resizebox{0.55\linewidth}{!}{
\begin{tabular}{l|cc|c|c|c}
\toprule[1.5pt]
\multirow{2}{*}{Method}   & \multicolumn{2}{c|}{Random Noises}       & \multirow{2}{*}{LAP} & \multirow{2}{*}{RAD} & \multirow{2}{*}{mIoU (\%)} \\ \cline{2-3}
                                                                               & \multicolumn{1}{c|}{Features} & Coordinates &                        &                         &                       \\ \midrule
Our Baseline                 &         &         &         &         & 50.1                 \\
Ours w/ Noise               & \checkmark &         &         &         & 49.1                 \\
Ours w/ Noise                &         & \checkmark &         &         & 52.9                 \\
Ours w/ Noise                 & \checkmark & \checkmark &         &         & 52.6                 \\ \midrule
Ours w/ PAP                  &         &         & \checkmark &         & 53.9                 \\
Ours w/ RAD                   &         &         &         & \checkmark & 54.8                 \\
Our DAT                 &         &         & \checkmark & \checkmark & \textbf{56.5}    \\  
\bottomrule[1.5pt]
\end{tabular}
}
\label{tab:s3dis-compare-baseline}
\end{table} 
Table~\ref{tab:s3dis-compare-sota} shows the results of our DAT and several SOTA methods on the S3DIS Area 5 dataset. Via effectively exploiting the unlabeled data, the DAT model with few labeled points training achieves comparable results with the upper bound (i.e. the fully-supervised KPConv model with 100\% labeled data training). Furthermore, under the ``OTOC'' setting, the DAT model significantly outperforms the second-best 1T1C method by 6.4\% mIOU gains on the S3DIS dataset. In addition, we further perform the ``One Thing Three clicks'' (OTTC) setting, where we annotate three points for each target. Our model outperforms the corresponding second-best method 1T1C~\cite{liu2021one} by 3.2\%.

\subsubsection{Ablation Studies}

\noindent\textbf{Comparisons with Baselines.}
We perform the ablation studies on the S3DIS dataset, to show the effectiveness of our proposed DAT. The first baseline is that we only use the segmentation loss $\mathcal{L}_{seg}$ on a few labeled points to train the segmentation model, which is denoted as ``Our Baseline'' in Table~\ref{tab:s3dis-compare-baseline}. Our proposed DAT outperforms ``Our baseline'' by 6.4\%. Another baseline is that we apply random noises to all the points to generate perturbed examples. Then we use KL-div loss to encourage the prediction consistency between the original point cloud and the perturbed examples. Specifically, similar to our designed LAP, we are able to apply random noises to point coordinates, point features, or both, which is denoted as ``Ours w/ Noise''. As Table~\ref{tab:s3dis-compare-baseline} shows, the DAT significantly outperforms two baseline methods, which suggests that our adaptive perturbations achieve better regularization to the unlabeled data compared to the random noises.

\noindent\textbf{Effects of LAP and RAD.}
To demonstrate the effects of two novel modules, as shown in Table~\ref{tab:s3dis-compare-baseline}, with separately applying consistency loss on the transformed examples generated by LAP (Ours w/ LAP) or RAD (Ours w/ RAD), we are able to significantly improve mIoU results compared with the ``Our Baseline''. This suggests that enforcing the consistency between the prediction of transformed examples and the original point clouds can predict better segmentation masks. ``Our DAT'' denotes that we apply the consistency loss on both LAP and RAD. Table~\ref{tab:s3dis-compare-baseline} shows combining both modules can further improve mIoU by 2.6\% and 1.7\% compared with only using LAP or RAD, respectively.

\begin{table}[t]
\centering
\caption{Ablation studies of our DAT about the Class-aware Perturbation Generator (CPG) used in our LAP module under the OTOC setting on the S3DIS dataset.}
\resizebox{0.6\linewidth}{!}{
\begin{tabular}{c|ccc|c|c}
\toprule[1.5pt]
\multirow{2}{*}{Method} & \multicolumn{3}{c|}{LAP}                                                                         & \multirow{2}{*}{RAD}      & \multirow{2}{*}{mIoU (\%)}          \\ \cline{2-4}
                        & \multicolumn{1}{c|}{Feat. w/o CPG} & \multicolumn{1}{c|}{Feat. w/ CPG} & Coordinates               &                           &                                \\ \midrule
Ours w/o RAD             & \checkmark         &                                  &                           &                           & 51.3                           \\
Ours w/o RAD             &                                   & \checkmark        &                           &                           & \textbf{51.7} \\ \midrule
Ours w/o RAD             & \checkmark         &                                  & \checkmark &                           & 53.3                           \\
Ours w/o RAD             &                                   & \checkmark        & \checkmark &                           & \textbf{53.9} \\ \midrule
Our DAT             & \checkmark         &                                  & \checkmark & \checkmark & 55.1                           \\
Our DAT             &                                   & \checkmark        & \checkmark & \checkmark & \textbf{56.5} \\ \bottomrule[1.5pt]
\end{tabular}
}
\label{tab:S3DIS-ablation-covar}
\end{table}
\noindent\textbf{Effects of CPG.}
We further verify the effectiveness of our designed CPG used in the LAP module. `` Feat. w/o CPG'' denotes that we generate the initial perturbation $d_f$ from the iid Gaussian distribution, instead of the class-aware multivariate Gaussian distribution. 
Table~\ref{tab:S3DIS-ablation-covar} shows that our class-aware perturbation generator is able to boost segmentation performance under all settings, which suggests that the class-aware information is critical in the point cloud segmentation task.

\begin{figure}[t]
\centering
    \includegraphics[width=0.8\linewidth]{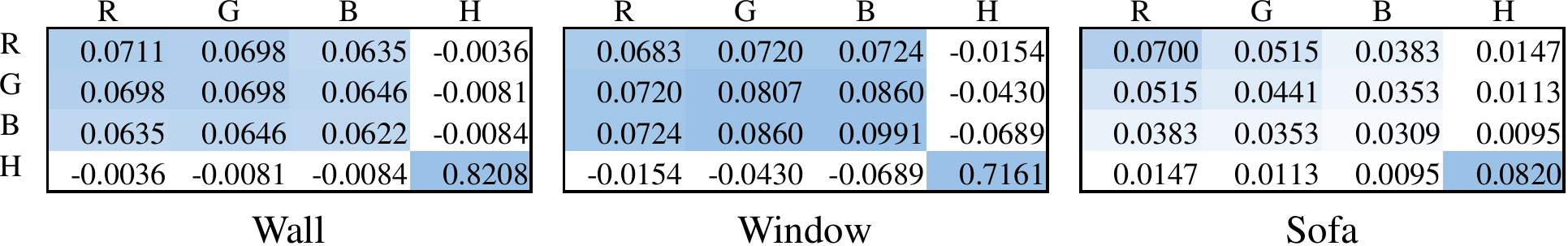}
    \caption{Three covariance matrices estimated via our designed CPG module under the OTOC setting on the S3DIS datasets.}
    \label{fig:covar}
\end{figure}

Besides, Figure~\ref{fig:covar} gives three examples of the computed covariance matrices in the CPG, where we randomly select them from all 13 covariance matrices. We can observe that different classes have different covariance matrices.

\begin{table}[t]
\centering
\caption{Ablation studies of our DAT on different affine transformations used in RAD module under the OTOC setting on the S3DIS dataset.}
\resizebox{0.5\linewidth}{!}{
\begin{tabular}{l|c|ccc|c}
\toprule[1.5pt]
\multirow{2}{*}{Method} & \multirow{2}{*}{LAP}    &                                  &              RAD              &                           & \multirow{2}{*}{mIoU (\%)} \\ \cline{3-5}
                        &                           & \multicolumn{1}{l|}{Translation} & \multicolumn{1}{l|}{Scale} & Rotation                  &                       \\ \hline
Ours w/ RAD          &                           & \checkmark        &                            &                           & 54.1   \\
Ours w/ RAD          &                           &                   &     \checkmark             &  & 54.6                   \\
Ours w/ RAD          &                           &                   &                            & \checkmark & 53.9                   \\
Ours w/ RAD          &                           & \checkmark        & \checkmark  &                           & \textbf{54.8}                  \\
Ours w/ RAD          &                           & \checkmark        &                            & \checkmark & 54.6                   \\
Ours w/ RAD          &                           & \checkmark        & \checkmark  & \checkmark                & 54.5            \\ \midrule
Our DAT         & \checkmark & \checkmark        &                            &                           & 55.5                  \\
Our DAT         & \checkmark &                   &   \checkmark               &                           & 55.9                  \\
Our DAT         & \checkmark &         &                            & \checkmark   & 55.2                  \\
Our DAT         & \checkmark & \checkmark        & \checkmark                 &                           & 56.0         \\
Our DAT         & \checkmark & \checkmark        &                            & \checkmark   & 55.2                  \\
Our DAT         & \checkmark & \checkmark        & \checkmark                 & \checkmark   & \textbf{56.5}                \\ \bottomrule[1.5pt]
\end{tabular}
}

\label{tab:S3DIS-ablation-affine}
\end{table}
\noindent\textbf{Different Affine Transformations in RAD.}
Table~\ref{tab:S3DIS-ablation-affine} shows the mIoU results for our DAT with different affine transformations. ``Ours w/ RAD'' indicates that we only apply the consistency loss on the deformed examples generated by RAD, and ``Our DAT'' indicates that we make use of all the transformed examples generated by LAP and RAD to train the model. As Table~\ref{tab:S3DIS-ablation-affine} shows, ``Our DAT'' achieves the best performance by using all three affine transformation methods (i.e. translation, scale and rotation).

\begin{table}[t]
\caption{To show the generalization ability, we further show the results with MinkowskiNet32 \cite{choy20194d} backbone on the S3DIS Area-5 set. ``Our DAT*'' denotes we only use our LAP module to train the backbone.}
\centering

\resizebox{0.55\linewidth}{!}{
\begin{tabular}{c|c|c}
\toprule[1.5pt]
Method                                                               & Supervision (\%) & mIoU (\%)                      \\ \midrule
Our Baseline                                                         & 0.02\% (OTOC)    & 48.7                           \\
Our Baseline                                                         & 0.06\% (OTTC)    & 55.0                           \\
Our DAT*                                                         & 0.02\% (OTOC)    & \textbf{54.6} \\
Our DAT*                                                         & 0.06\% (OTTC)    & \textbf{58.2} \\ \midrule
Our Upper bound                                                      & 100\%            & 65.4                           \\ \bottomrule[1.5pt]
\end{tabular}
}
\label{tab:s3dis-ablation-mink}
\end{table}
\begin{table}
\centering
\caption{Comparison of our DAT model with several existing methods on the ScanNet-v2 test set. `` Our DAT$\dagger$'' denotes that our DAT is built upon the 1T1C~\cite{liu2021one} model.}
\resizebox{0.55\linewidth}{!}{
  \begin{tabular}{c|c|c}
    \toprule[1.5pt]
    Method & Supervision & mIoU (\%)  \\
    \midrule
    Pointnet++~\cite{qi2017pointnet++} & 100\% &33.9 \\
    PointCNN~\cite{li2018pointcnn} & 100\% & 45.8\\
    MinkowskiNet~\cite{choy20194d} &100\% & 73.6 \\
    Virtual MVFusion~\cite{kundu2020virtual} &100\%+2D & 74.6\\
    MPRM~\cite{wei2020multi} & scene-level & 24.4 \\    
    MPRM~\cite{wei2020multi} & subcloud-level & 41.1 \\    
    MPRM+CRF~\cite{wei2020multi} & subcloud-level & 43.2 \\  
    
    \midrule \midrule
    CSC\_LA\_SEM~\cite{hou2021exploring} & 20 points & 53.1 \\
    Viewpoint\_BN\_LA\_AIR~\cite{luo2021pointly}	& 20 points & 54.8 \\
    PointContrast\_LA\_SEM~\cite{xie2020pointcontrast} & 20 points & 55.0 \\
    1T1C~\cite{liu2021one} & 20 points & 59.4\\
    \midrule
    Our Baseline & 20 points & 51.6\\
    Our DAT & 20 points & 55.2\\
    Our DAT$\dagger$ & 20 points & \textbf{62.3}\\
    \midrule
    Our Upper Bound  & 100\% & 68.4 \\
    
    \bottomrule[1.5pt]
  \end{tabular}
}

\label{tab:scannet-test}
\end{table}

\begin{table}
\centering
\caption{Comparison of our DAT model with several existing methods on the ScanNet-v2 validation set. `` Our DAT$\dagger$'' denotes that our DAT is built upon the 1T1C~\cite{liu2021one} model.} 
\resizebox{0.45\linewidth}{!}{
  \begin{tabular}{c|c|c}
    \toprule[1.5pt]
    Method & Supervision & mIoU (\%) \\
    \midrule
    1T1C~\cite{liu2021one} & 20 points & 61.4 \\ \midrule
    Our Baseline  & 20 points & 54.6\\
    Our DAT & 20 points & {58.9}\\ 
    Our DAT$\dagger$ & 20 points & \textbf{64.4}\\ \midrule
    Our Upper Bound   & 100\% & 68.5 \\
    \bottomrule[1.5pt]
  \end{tabular}
}
\label{tab:scannet-val}
\end{table}
\subsubsection{Generalization Ability.}
To verify the generalization ability, we further use our training strategy to train a voxel-based segmentation framework (i.e. MinkowskiNet \cite{choy20194d}). Unlike the point-based methods, the voxel-based methods firstly project the point cloud into regular voxels and then apply 3D sparse convolution on it. Since the projecting operation is non-differentiable and cannot back-propagate the gradients to point coordinates, we only employ the LAP module to add adaptive perturbations on the input features with the CPG module (labeled as ``Our DAT*'' in Table~\ref{tab:s3dis-ablation-mink}). Table~\ref{tab:s3dis-ablation-mink} shows, under the OTOC/OTTC setting, our model improves the mIoU results by 5.9\%/3.2\% compared to their respective ``Our Baseline'', which demonstrates that such novel training strategy is general and effective, and can be easily applied to various point cloud frameworks.

\subsection{Evaluations on ScanNet-v2 dataset}

\begin{figure}[t]
\centering
    \includegraphics[width=0.9\linewidth]{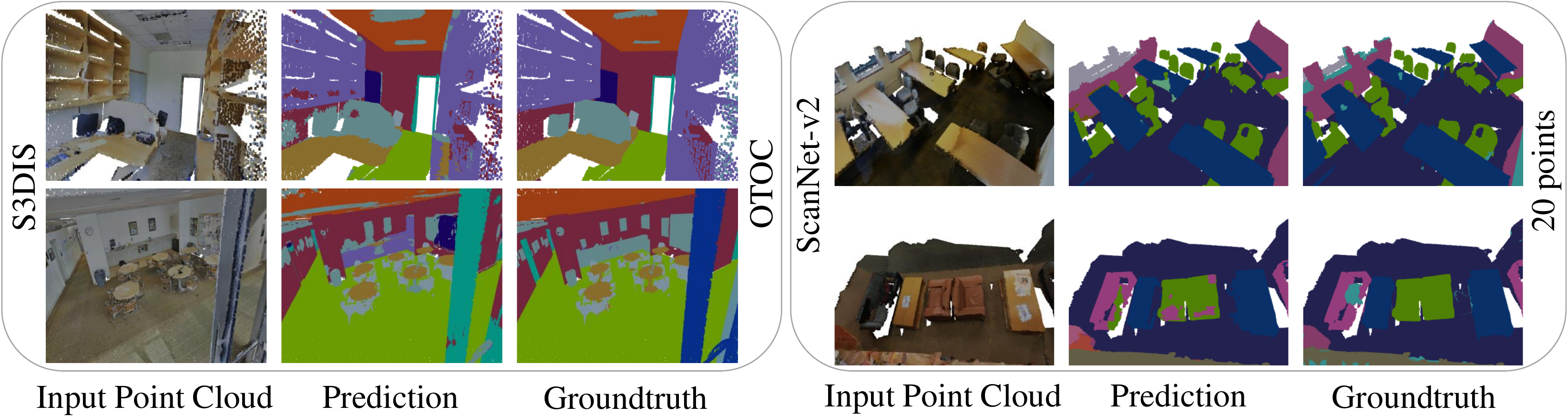}
    \caption{Two results of our DAT on the S3DIS (first two rows, under the ``OTOC'' setting) and ScanNet-v2 datasets (last two rows, under the ``20 points'' setting).}
    \label{fig:vislize_result}
\end{figure}
Tables~\ref{tab:scannet-test} and \ref{tab:scannet-val} respectively give the results on the test and validation set of ScanNet-v2 dataset in the ``3D Semantic label with Limited Annotations'' benchmark. We use the officially given 20 points annotations as the sparse labels to train the model. Compared with ``Our Baseline'', our DAT (denoted as ``Our DAT'') with the KPConv backbone can achieve impressive performance gains of 3.2\% and 3.9\% mIoU on ScanNet-v2 test and validation sets, respectively. 

Meanwhile, such a training strategy can be easily combined with existing models for point cloud segmentation. For example, on the ScanNet-v2 dataset, we build our DAT upon the 1T1C model, which is used to generate the pseudo labels for all training data. Then we use the pseudo labels to train our DAT. Based on the 1T1C model (denoted as ``Our DAT $\dagger$'' in Tables~\ref{tab:scannet-test} and \ref{tab:scannet-val}), our DAT can further improve the mIoU results by 2.9\% and 3.0\% on the ScanNet-v2 test and validation set compared with 1T1C, respectively. This suggests that our training strategy can further improve the performance of other SOTA models.

\subsection{Qualitative Results}
Figure~\ref{fig:vislize_result} shows the segmentation results obtained by our proposed DAT model on the S3DIS and ScanNet-v2 dataset. It reveals that the DAT model can successfully preserve most of the object structures and segment the 3D point clouds accurately, only with the weak annotation training.

\section{Conclusion}
In this paper, we have presented a Dual Adaptive Transformations (DAT) model for the weakly supervised point cloud segmentation task, with two novel designs, i.e. the LAP and RAD module. First, the LAP module generates point-wise adaptive coordinate perturbations and class-aware adaptive feature perturbations based on the online estimated class distribution. Second, we propose the RAD module to generate regional adaptive deformations by applying a set of adaptive affine transformations on the superpoint regions. Extensive experimental results under multiple weakly supervised settings have demonstrated that our proposed DAT model achieves new SOTA segmentation performance on the S3DIS and ScanNet-v2 datasets. 

\noindent\textbf{Acknowledgments.} This study is supported under the RIE2020 Industry Alignment Fund – Industry Collaboration Projects (IAF-ICP) Funding Initiative, as well as cash and in-kind contribution from the industry partner(s). This research is partly supported by the National Research Foundation, Singapore under its AI Singapore Programme (AISG Award No: AISG-RP-2018-003), the Ministry of Education, Singapore, under its Academic Research Fund Tier 2 (MOE-T2EP20220-0007) and Tier 1 (RG95/20). This research is also partially supported by Monash FIT Start-up Grant and SenseTime Gift Fund.

%
%
\bibliographystyle{splncs04}
\bibliography{egbib}
\clearpage
\begin{center}
\textbf{\Large Dual Adaptive Transformations for Weakly Supervised Point Cloud Segmentation} \\[5pt]

\textbf{\Large Supplementary Material} \\

\end{center}

\renewcommand{\thesection}{\Alph{section}}
\renewcommand{\thetable}{\Alph{table}}
\renewcommand{\thefigure}{\Alph{figure}}
\renewcommand{\thealgorithm}{\Alph{algorithm}}

\setcounter{section}{0}
\setcounter{table}{0}
\setcounter{figure}{0}
\setcounter{algorithm}{0}

\section{More Ablation Studies}

\noindent\textbf{Different hyper-parameters $\alpha$ and $\beta$.}
We conduct experiments with different $\alpha$ and $\beta$ under the OTOC setting on the S3DIS dataset. From Table~\ref{tab:ablation_hyper}, we can see that setting both $\alpha$ and $\beta$ as 2 yields the best performance for the weakly supervised point cloud segmentation task.

\begin{table*}[h]
\caption{Ablation studies on different $\alpha$ and $\beta$ under the OTOC setting on the S3DIS dataset.}
\centering
		\begin{minipage}[t]{0.3\textwidth}
		\centering
			\setlength{\tabcolsep}{5pt}
			\begin{center}
				\begin{tabular}{c|c|c}
\toprule[1.5pt]
$\alpha$                & $\beta$                  & mIoU(\%) \\ \midrule
\multirow{3}{*}{2} & 1                  &   55.6    \\
                     & 2                   &  \textbf{56.5}    \\
                     & 5                   & 56.2      \\ \bottomrule[1.5pt]
\end{tabular}
			\end{center}
			\begin{center}

			\end{center}
		\end{minipage}
		\begin{minipage}[t]{0.3\textwidth}
		\centering
			\setlength{\tabcolsep}{5pt}
			\begin{center}
				\begin{tabular}{c|c|c}
                \toprule[1.5pt]
                $\alpha$                & $\beta$                 & mIoU(\%) \\ \midrule
                1                  & \multirow{3}{*}{2} &  55.1    \\
                2                  &                       &  \textbf{56.5}    \\
                5                  &                       &   56.1    \\ \bottomrule[1.5pt]
                \end{tabular}
			\end{center}
			\begin{center}

			\end{center}
		\end{minipage}
\label{tab:ablation_hyper}
\end{table*}

\noindent\textbf{Different Superpoint in RAD.} We conduct experiments to analyze the sensitivity of superpoints, which are used in RAD. Specifically, we generate the superpoints with different hyperparameters of the number of neighbors for the geometric features $geo$ and adjacency graph $adj$. As shown in Table~\ref{tab:ablation_superpoint}, we can see that our RAD module is not very sensitive to different superpoints.

\begin{table*}[h]
\caption{Ablation studies on different $geof$ and $adj$ under the OTOC setting on the S3DIS dataset.}
\centering
		\begin{minipage}[t]{0.3\textwidth}
		\centering
			\setlength{\tabcolsep}{5pt}
			\begin{center}
				\begin{tabular}{c|c|c}
\toprule[1.5pt]
$geof$                & $adj$                  & mIoU(\%) \\ \midrule
30                     & 10                   & 54.4     \\
45                     & 5                   & 54.7      \\
45                     & 10                   & \textbf{56.5}      \\
\bottomrule[1.5pt]
\end{tabular}
			\end{center}
			\begin{center}

			\end{center}
		\end{minipage}

\label{tab:ablation_superpoint}
\end{table*}

\section{Psudeocodes}

Algorithm~\ref{alg:pap} and \ref{alg:rad} respectively provide the pseudocodes of our designed LAP and RAD modules.

\section{More Qualitative Results}

\begin{figure}[h]
\centering
    \includegraphics[width=0.6\linewidth]{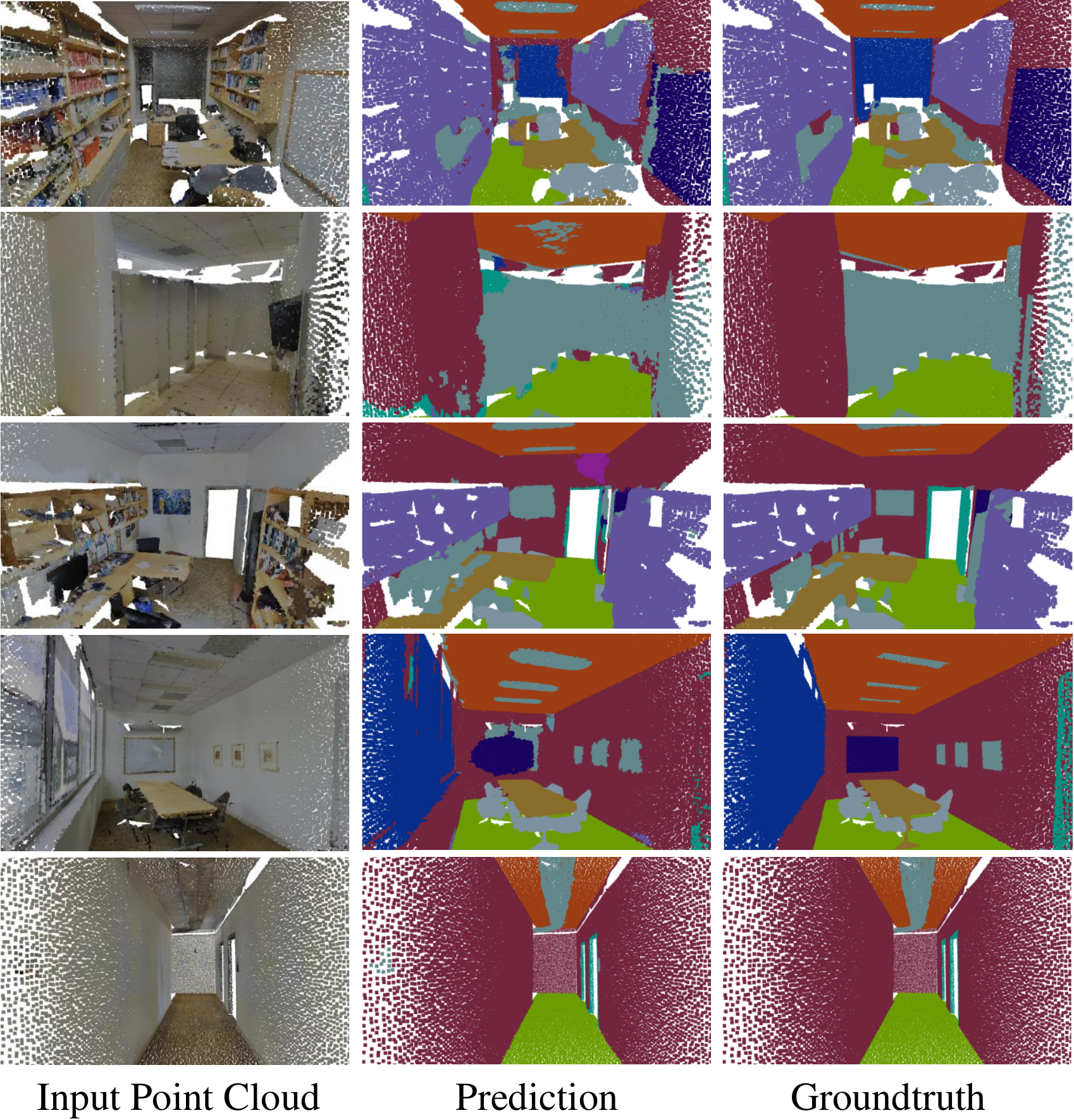}
    \caption{Visualization results obtained by our DAT model on the S3DIS dataset under the ``OTOC'' setting.}
    \label{fig:supp_s3dis}
\end{figure}

\begin{figure}[h]
\centering
    \includegraphics[width=0.6\linewidth]{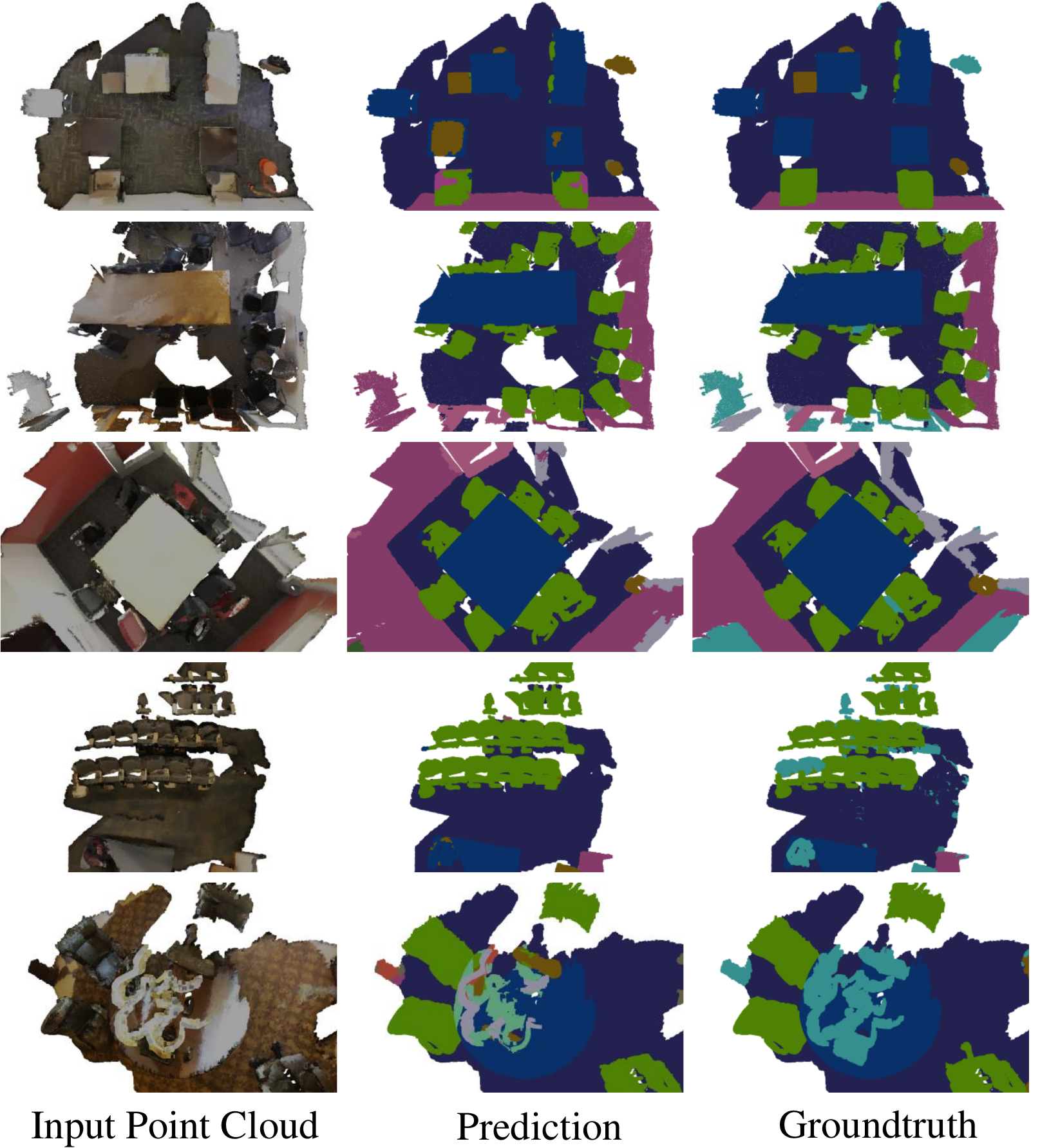}
    \caption{Visualization results obtained by our DAT model on the ScanNet-v2 dataset under the ``20 points'' setting.}
    \label{fig:supp_scannet}
\end{figure}

Figure~\ref{fig:supp_s3dis} and \ref{fig:supp_scannet} respectively show five more segmentation results obtained by our proposed model on the S3DIS and ScanNet-v2 dataset. Our DAT model is able to generate accurate segmentation masks for most of the points only with the weak annotation training.

\begin{algorithm*}[h]
\caption{Pseudocode of Local Adaptive Perturbation (LAP) module in a PyTorch-like style.}
\label{alg:pap}
\definecolor{codeblue}{rgb}{0.25,0.5,0.5}
\lstset{
  backgroundcolor=\color{white},
  basicstyle=\fontsize{7.2pt}{7.2pt}\ttfamily\selectfont,
  columns=fullflexible,
  breaklines=true,
  captionpos=b,
  commentstyle=\fontsize{7.2pt}{7.2pt}\color{codeblue},
  keywordstyle=\fontsize{7.2pt}{7.2pt},
}
\begin{lstlisting}[language=python]
# CPG: class-aware perturbation generator
# c: point coordinates
# f: point features
# xi_c: theta for coordinates
# xi_f: theta for features
# eps_c: epsilon for coordinates
# eps_f: epsilon for features
# ip: iteration times of computing adv noise (default: 1)

import torch.nn.functional as F

def LAP(c, f, model):

    pred = model(c, f)
    
    pseudo_label = pred.max(dim=1)[1] # generate pseudo label
    CPG.update_CV(c, f, pseudo_label) # update the covariance matrices
    
    c_init, f_init = CPG.generator(c, f, pseudo_label) # generate initial unit vectors for coordinates and features
    
    # normalize the initial unit vectors
    c_init = _l2_normalize(c_init)
    f_init = _l2_normalize(f_init)
    
    for _ in range(ip):
        c_init.require_grad()
        f_init.require_grad()
    
        pred_init = model(c + xi_c * c_init, f + xi_f * f_init)
        adv_distance = F.kl_div(F.log_softmax(pred_init, dim=1), pred)
        
        # generate adversarial perturbations
        adv_distance.backward()
        
        c_adv = _l2_normalize(c_init.grad) 
        f_adv = _l2_normalize(f_init.grad)
        
        model.zero_grad()
    
    pred_hat = model(c + eps_c * c_adv, f + eps_f * f_adv)
    
    Loss_pc = F.kl_div(F.log_softmax(pred_hat, dim=1), pred) # point-level consistency loss
    
    
\end{lstlisting}
\end{algorithm*}

\begin{algorithm*}[h]
\caption{Pseudocode of Regional Adaptive Deformation (RAD) module in a PyTorch-like style.}
\label{alg:rad}
\definecolor{codeblue}{rgb}{0.25,0.5,0.5}
\lstset{
  backgroundcolor=\color{white},
  basicstyle=\fontsize{7.2pt}{7.2pt}\ttfamily\selectfont,
  columns=fullflexible,
  breaklines=true,
  captionpos=b,
  commentstyle=\fontsize{7.2pt}{7.2pt}\color{codeblue},
  keywordstyle=\fontsize{7.2pt}{7.2pt},
}
\begin{lstlisting}[language=python]
# c: point coordinates
# f: point features
# xi: theta
# eps: epsilon
# ip: iteration times of computing adv noise (default: 1)

import torch.nn.functional as F

def RAD(c, f, model):

    pred = model(c, f)
    
    S = SP_G(c, f) # offline superpoint extraction
    
    A_init = A_generator(c, S) # generate initial affine transformation matrices
    A_init = normalize(A_init) # normalize the matrices
    
    for _ in range(ip):
        A_init.require_grad()
        
        c_init = affine(c, S, A_init) # generate initial perturbed point cloud
        
        pred_init = model(c_init, f) 
        
        adv_distance = F.kl_div(F.log_softmax(pred_init, dim=1), pred)
        
        # generate region-level adversarial examples
        adv_distance.backward() 
        A_adv = normalize(A_init.grad)
        model.zero_grad()

    c_adv = affine(c, S, eps * A_adv) 
    pred_hat = model(c_adv, f)
    
    Loss_rc = F.kl_div(F.log_softmax(pred_hat, dim=1), pred) # region-level consistency loss
    
    
\end{lstlisting}
\end{algorithm*}

\end{document}